\documentclass[conference]{IEEEtran}
\IEEEoverridecommandlockouts

\usepackage{cite}
\usepackage{amsmath,amssymb,amsfonts}
\usepackage{algorithmic}
\usepackage{graphicx}
\usepackage{textcomp}
\usepackage{xcolor}
\usepackage{booktabs}
\usepackage{url}
\usepackage{hyperref}
\usepackage[utf8]{inputenc}
\usepackage{newunicodechar}
\newunicodechar{−}{-}

\def\BibTeX{{\rm B\kern-.05em{\sc i\kern-.025em b}\kern-.08em
    T\kern-.1667em\lower.7ex\hbox{E}\kern-.125emX}}

\begin{document}

\title{Vision-Language Grounded Semantic Navigation for Autonomous Mobile Robots}

\author{
    \IEEEauthorblockN{Ravindran Vasoor Saravanan\textsuperscript{1}, Chittin Varunav Ravi\textsuperscript{1}, Thirulok Sundar Mohan Rasu\textsuperscript{2}, Madhumita Singh\textsuperscript{1}}
    \IEEEauthorblockA{
        \textsuperscript{1}Department of Mechanical Engineering, \textsuperscript{2}Department of Electrical Engineering and Computer Science \\
        University of Michigan, Ann Arbor, MI, USA \\
        \textbf{GitHub Repository:} \href{https://github.com/Ravindran127/Vision-Language-Grounded-Semantic-Navigation-for-Autonomous-Mobile-Robots}
     {VLG SN for Autonomous Mobile Robots} \\
    }
    }

\maketitle

\begin{abstract}
Language-guided navigation enables mobile robots to interpret natural language commands and navigate toward user-specified targets in unstructured environments. However, integrating open-vocabulary visual perception with classical robotics pipelines for mapping and navigation remains a challenging problem. In this work, we present a complete ROS2-based system that combines frontier-based exploration, open-vocabulary object detection using Grounding DINO, depth-based 3D localization, and semantic mapping to enable end-to-end vision-language grounded navigation in simulation.
The system accepts natural language commands such as ``go to the yellow box,'' detects candidate objects from RGB images using text prompts, estimates their positions in the world frame, and generates navigation goals executed by the Nav2 stack. In a Gazebo-based indoor environment with a TurtleBot3 robot, the system autonomously explored the room, detected three of the five target objects, and successfully executed both natural language navigation commands tested, without task-specific training. The results indicate that open-vocabulary grounding models can be integrated into classical navigation pipelines, though performance is limited by detection uncertainty, depth-based localization error, and incomplete exploration coverage. These findings highlight both the feasibility and current limitations of deploying vision-language grounded navigation systems using lightweight, CPU-only architectures.
\end{abstract}


%-----------------------------------------------------------------------
\section{Introduction}

Autonomous mobile robots in human environments must do more than move from point A to point B: they must understand their surroundings in terms that humans naturally use. When a user says ``go to the blue box'' or ``find the red cylinder,'' the robot needs to understand both the spatial layout of the environment and the semantic properties of the objects within it. This problem, known as vision-language grounded navigation, combines simultaneous localization and mapping (SLAM), open-vocabulary computer vision, and natural language processing (NLP).

Traditional navigation systems model the world as a geometric entity consisting of walls, obstacles, and free space. Semantic mapping extends this by adding labels to objects, but most systems can only recognize a fixed set of object types learned during training~\cite{sunderhauf2018}. This leads to two key limitations. First, adapting a closed-vocabulary system to a new domain requires collecting labeled data and retraining the detector. Second, the system can only recognize objects it was trained on, making it difficult to handle new or unseen object descriptions at runtime.

Recent vision-language models, especially Grounding DINO~\cite{liu2023groundingdino}, address this by allowing open-vocabulary object detection using text prompts. Users can describe objects in natural language, and the model attempts to find them in images. When combined with language-driven planning systems such as SayPlan~\cite{rana2023sayplan} and LM-Nav~\cite{shah2023lmnav}, these capabilities open the door to robots that can be instructed in ordinary English.

However, integrating these models with standard navigation systems is still challenging. Grounding DINO is slow on CPU (about 5--10 seconds per image), detections can be noisy when multiple objects are present, and keeping detections consistent with the SLAM map and user commands requires careful system design. In this work, we focus on addressing these challenges within a unified system.

\subsection{Contributions}

This work presents an integrated system that combines open-vocabulary perception with classical robotics components. Our main contributions are:

\begin{enumerate}
    \item \textbf{ROS2-based integrated pipeline:} We implement a ROS2 Humble--based system combining frontier-based exploration, Grounding DINO object detection, depth-based 3D localization, semantic mapping, and command parsing. These components are coordinated through a state machine managing transitions between exploration, perception, and navigation.

    \item \textbf{Frontier selection with failure handling:} We implement a frontier selection strategy that scores candidate frontiers based on size and distance. The system tracks repeated navigation failures and adjusts behavior to reduce the likelihood of getting stuck in difficult regions, such as concave areas or unreachable goals.

    \item \textbf{Rule-based natural language command parsing:} We implement a simple regex-based parser to extract key elements such as action, object category, and visual attributes (e.g., color), enabling consistent interpretation of navigation commands.

    \item \textbf{Experimental validation in simulation:} The system runs on Ubuntu 22.04 with ROS2 Humble using CPU-only PyTorch, without requiring GPU acceleration. We demonstrate end-to-end operation from autonomous exploration to language-guided navigation in a Gazebo-based indoor environment with geometric objects.
\end{enumerate}

\begin{figure}[htbp]
    \centering
    \includegraphics[width=\columnwidth]{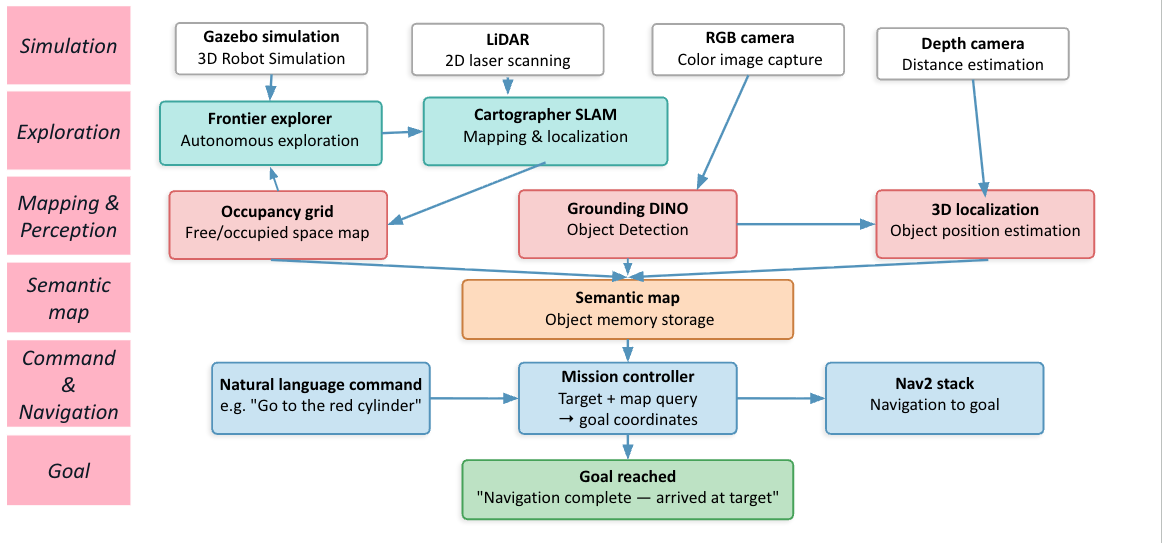}
    \caption{System architecture of the vision-language grounded navigation pipeline, showing the five-layer structure from Gazebo simulation through sensor processing, perception, planning, and mission control.}
    \label{fig:architecture}
\end{figure}

The complete source code for this system is publicly available as open-source software on GitHub (see repository link on the title page).

\subsection{System Overview}

The system operates in two main phases. During the \emph{exploration phase}, the robot autonomously navigates an unknown environment using a frontier-based strategy. At regular intervals (approximately every 1.5 meters), Grounding DINO processes the current camera image to detect objects, which are then localized in 3D using depth-based projection and added to a semantic map. During the \emph{query phase}, the user issues natural language commands that are parsed and matched against the semantic map to generate navigation goals, which are executed by the Nav2 stack.

%-----------------------------------------------------------------------
\section{Related Work}

\subsection{Frontier-Based Exploration}

The frontier exploration algorithm, originally proposed by Yamauchi~\cite{yamauchi1997}, defines frontiers as boundaries between known free space and unknown space in an occupancy grid. The robot iteratively navigates to frontier centroids, expanding the mapped region until no frontiers remain. This foundational approach has been extended with methods that use information gain~\cite{stachniss2005}, coordinated multi-robot exploration to reduce overlap~\cite{burgard2005}, and strategies that balance exploration benefit with navigation cost.

Our work builds on the classical Yamauchi approach with two modifications. First, we use a weighted scoring function that balances frontier size (as a proxy for information gain) against Euclidean distance from the robot, favoring large nearby frontiers. Second, we implement an adaptive failure-escalation policy that switches between selection strategies based on consecutive navigation failures, preventing the robot from becoming permanently stuck.

\subsection{Semantic Mapping}

Semantic mapping systems associate semantic labels with geometric map representations. Early approaches relied on convolutional neural networks trained on fixed sets of object categories. More sophisticated systems like SemanticFusion~\cite{mccormac2017} fuse convolutional neural network predictions with dense SLAM to produce 3D semantic reconstructions. Recent work on ConceptFusion~\cite{jatavallabhula2023} demonstrates open-vocabulary 3D mapping using multimodal foundation models, allowing arbitrary text queries against the reconstructed scene.

Our approach differs in two ways. First, we use Grounding DINO's open-vocabulary 2D detection combined with depth-image projection, rather than dense 3D reconstruction, which significantly reduces computational cost. Second, our semantic map is lightweight (a list of labeled 3D points with confidence scores and observation counts), making it easy to store in JSON and query using simple string matching.

\subsection{Vision-Language Models for Robotics}

The integration of large vision-language models into robotics has gained significant momentum. CLIP~\cite{radford2021} enabled zero-shot visual recognition from text descriptions by training on 400 million image-text pairs. Grounding DINO~\cite{liu2023groundingdino} brings open-vocabulary recognition to object detection: it fuses language into the transformer-based DINO detector through grounded pre-training, so that it predicts bounding boxes for arbitrary text phrases. SayPlan~\cite{rana2023sayplan} and LM-Nav~\cite{shah2023lmnav} demonstrate language-driven navigation using large language models for task planning and landmark grounding.

Our approach takes a simpler direction. Rather than using large language models for runtime planning, we use Grounding DINO only for perception and process natural language commands with regex-based parsing. This makes the system deterministic, fast to debug, and deployable without API dependencies or GPU requirements.

\subsection{Object-Goal Navigation}

Object-goal navigation refers to the task of navigating to an instance of a specified object category in an unknown environment~\cite{batra2020}. Methods for this problem range from modular systems that build semantic maps and use classical planning~\cite{chaplot2020} to end-to-end reinforcement learning approaches that learn navigation policies directly from visual input. Our system considers a related setting where targets are specified using natural language descriptions, including attributes such as color. The robot first explores the environment to build a semantic map and then uses it to navigate to the requested object.

%-----------------------------------------------------------------------
\section{System Architecture}
\label{sec:arch}

\subsection{Simulation Environment}

Experiments are conducted in Gazebo Classic 11 using a TurtleBot3 Waffle Pi differential-drive robot. The robot is equipped with (i) a 360$^\circ$ HLS-LFCD LiDAR (0.12--3.5\,m range), (ii) an RGB camera ($640\times480$, 62.2$^\circ$ horizontal field of view), and (iii) a co-located depth camera publishing 32-bit float depth images. The simulation runs in real time using ODE physics with a 0.001\,s step size.

The environment consists of an 8\,m\,$\times$\,8\,m indoor room enclosed by 2.5\,m walls. Five geometric objects are placed at fixed locations: red cylinder $(-2,\,2)$, blue box $(2,\,2)$, green cylinder $(-2,\,-2)$, yellow box $(2.5,\,-2.5)$, and white cylinder $(2.5,\,0.5)$. Objects use distinct solid colors to ensure clear visual separation. The robot is initialized at the origin.

\begin{figure}[htbp]
    \centering
    \includegraphics[width=\columnwidth]{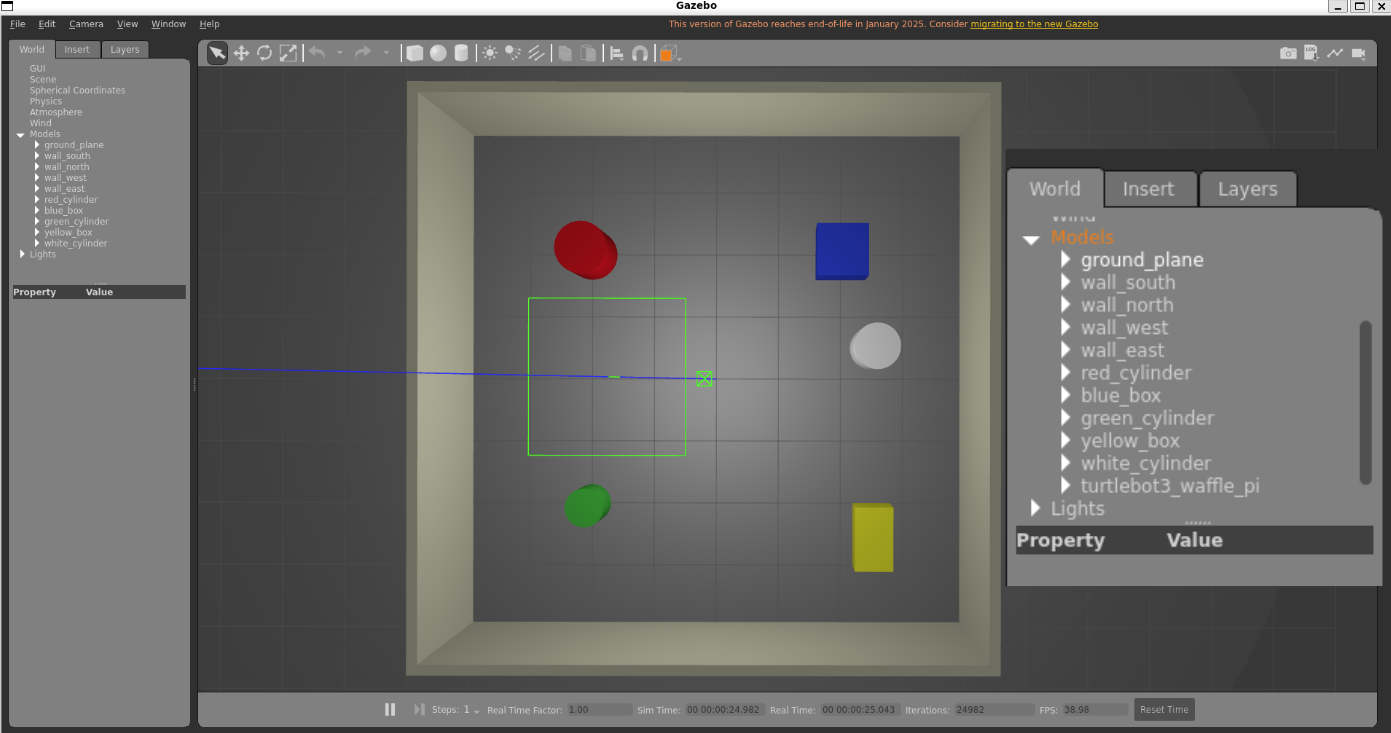}
    \caption{Gazebo simulation environment containing five geometric objects used for evaluation.}
    \label{fig:gazebo}
\end{figure}

Cartographer SLAM~\cite{hess2016} processes LiDAR data to generate a 2D occupancy grid at 0.05\,m resolution. The map is published at 1\,Hz and used by both the frontier exploration module and the Nav2~\cite{macenski2020} navigation stack.

\subsection{Frontier-Based Exploration}

The \texttt{FrontierExplorer} module processes the 2D occupancy grid in three steps. First, each cell is classified as \emph{free} (occupancy value $<50$), \emph{occupied} ($\geq 50$), or \emph{unknown} ($-1$). Second, the unknown region is dilated by one pixel and intersected with the free mask to obtain frontier cells---free cells adjacent to unknown space. Third, connected-component labeling clusters frontier cells into frontier regions, filtering out clusters smaller than $k_{\min} = 3$ cells.

For frontier selection, we implement and compare four strategies:
\begin{align}
\text{Closest:}  &\quad \arg\min_i\; d(r,\, c_i) \notag\\
\text{Largest:}  &\quad \arg\max_i\; |F_i| \notag\\
\text{Farthest:} &\quad \arg\max_i\; d(r,\, c_i) \notag\\
\text{Weighted:} &\quad \arg\max_i\; \frac{|F_i|}{d(r,\, c_i)+1} \notag
\end{align}
where $|F_i|$ is the size of frontier cluster $i$, $c_i$ is its centroid, and $d(r,\,c_i)$ is the Euclidean distance from the robot's grid position $r$. The weighted strategy balances information gain (larger frontiers are more valuable) against travel efficiency.

\textbf{Adaptive failure recovery.} An important addition to our frontier explorer is the failure-escalation policy. When Nav2 reports consecutive navigation failures (goal rejected, timed out, or canceled because the robot is physically stuck), the system escalates through four levels:
{\renewcommand{\labelenumi}{(\alph{enumi})}\begin{enumerate}
    \item Up to 1 failure: use the configured strategy (default weighted).
    \item 2--3 failures: switch to \emph{farthest} to escape concave regions.
    \item 4--5 failures: clear the visited-frontier list and use \emph{farthest}.
    \item 6+ failures: full reset, resume normal strategy.
\end{enumerate}}
A simple stuck-detection mechanism is used to identify when the robot is not making progress. If the robot moves less than a small threshold (0.3\,m) over a fixed time interval (45\,s), the system triggers a recovery or re-selection of exploration targets.

\begin{figure}[htbp]
    \centering
    \includegraphics[width=0.6\columnwidth]{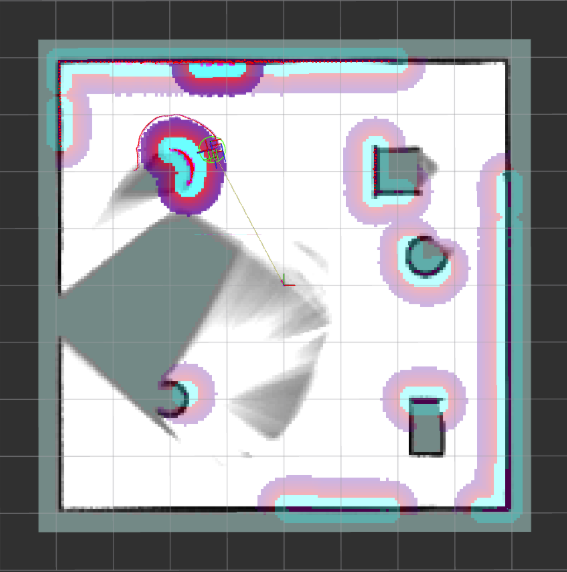}
    \caption{Frontier detection on the Cartographer occupancy grid. White cells denote known free space, black cells denote walls/obstacles, and gray cells denote unknown regions. Frontier cells (the boundary between free and unknown space) are clustered, and the weighted strategy selects the navigation goal shown.}
    \label{fig:frontier}
\end{figure}

\subsection{Object Detection with Grounding DINO}

The \texttt{GroundingDINODetector} module wraps the HuggingFace implementation of Grounding DINO (model ID: \texttt{IDEA-Research/grounding-dino-base}, approximately 232\,M parameters). The model weights are downloaded once and cached locally for subsequent use. Inference takes approximately 5--10 seconds per image on CPU, which is acceptable for the 1.5-meter detection interval but not suitable for real-time operation without GPU acceleration.

Detection is triggered during exploration whenever the robot has traveled more than 1.5 meters since the last detection. The detection prompt consists of a set of candidate object descriptions tailored to the simulation environment.

Grounding DINO returns a list of bounding boxes with per-box confidence scores and text labels. A minimum confidence threshold of 0.28 (empirically tuned) filters spurious detections. Detections are sorted by descending confidence before being passed to the 3D localization stage.

\textbf{Label cleaning.} A subtle issue with Grounding DINO is that when multiple objects appear in a single image frame, the detector sometimes concatenates multiple prompt tokens into a single label (e.g., ``red cylinder blue box green cylinder''). Our label cleaning heuristic matches the detected label against all prompt terms and returns the longest matching single term. This partially resolves the multi-object label concatenation problem but is not a complete solution.

\subsection{3D Object Localization from 2D Bounding Boxes}
\label{sec:localization}

The \texttt{BBoxTo3DConverter} module converts 2D bounding box detections to 3D world coordinates through a three-step pipeline.

\textbf{Step 1: Depth extraction.} For each bounding box $[x_1, y_1, x_2, y_2]$, we compute the median depth value from the central 50\% region of the bounding box in the aligned depth image. We filter depth values outside $[0.1,\, 10.0]$\,m to remove invalid readings.

\textbf{Step 2: Pixel-to-camera projection.} The bounding box center $(u,\,v)$ is unprojected using the pinhole camera model:
\begin{equation}
\mathbf{p}_c = \left[\frac{(u-c_x)\,d}{f_x},\; \frac{(v-c_y)\,d}{f_y},\; d\right]^\top
\end{equation}
where $d$ is the extracted depth and $(f_x,\,f_y,\,c_x,\,c_y)$ are camera intrinsics with $c_x = 320$, $c_y = 240$, and $f_x = f_y$ computed from the image width $w$ and horizontal field of view $\theta$ via $f_x = w / (2\tan(\theta/2))$. We note a discrepancy: the Gazebo SDF defines HFOV\,$=$\,1.086\,rad (62.2$^\circ$), yielding $f_x = 530.5$, whereas our code uses a hard-coded default of 69.4$^\circ$, yielding $f_x = 462.1$. This $\sim$13\% focal-length mismatch contributes to 3D localization error.

\textbf{Step 3: Camera-to-world transform.} The 3D point is transformed to the world frame using the camera-to-world homogeneous transform $T_{cw}$ retrieved from the ROS2 TF tree:
\begin{equation}
\begin{bmatrix} \mathbf{p}_w \\ 1 \end{bmatrix} = T_{cw} \begin{bmatrix} \mathbf{p}_c \\ 1 \end{bmatrix}
\end{equation}
If the TF lookup fails (for example, during initialization), the system falls back to using the odometry pose combined with a fixed camera-to-robot transform.

\subsection{Semantic Map}

The \texttt{SemanticMap} module maintains a list of \texttt{SemanticObject} entries, each containing a label (string), a 3D world position (list of 3 floats), a confidence score, a timestamp, and an observation count. When a new detection arrives, the system first checks for existing objects with the same label within a merge distance threshold $d_{\text{merge}} = 0.5$\,m. If a match is found, the existing object's position is updated via weighted running average:
\begin{equation}
\mathbf{p}_{\text{new}} = \frac{n\,\mathbf{p}_{\text{old}} + \mathbf{p}_{\text{det}}}{n+1}
\end{equation}
where $n$ is the existing observation count. The confidence is updated to the maximum of the old and new values, and the observation count is incremented.

The map supports several query types: fuzzy substring (case-insensitive substring matching, so ``blue'' matches ``blue box''), exact match, nearest to position (within a maximum distance), and spatial region (all objects within a spherical radius). The map serializes to JSON for persistence across runs, enabling incremental map building over multiple exploration sessions.

\subsection{Natural Language Command Parser}

The \texttt{CommandParser} module processes natural language navigation commands using a cascade of regular expressions. This design choice is deliberate: it avoids large language model inference, makes the parser deterministic and easy to debug, and eliminates external API dependencies. The parsing pipeline proceeds through four stages:

\begin{enumerate}
    \item \textbf{Action detection:} Patterns match action verbs such as \emph{go to}, \emph{navigate to}, \emph{move to}, \emph{find}, \emph{locate}, \emph{where is}, and \emph{explore}.
    \item \textbf{Spatial relation extraction:} Patterns match spatial prepositions (e.g., \emph{near}, \emph{behind}) followed by a reference object name.
    \item \textbf{Attribute extraction:} A color modifier is extracted from a vocabulary of five colors (red, blue, green, yellow, white).
    \item \textbf{Target extraction:} The text remaining after the above elements are removed becomes the target object label; in our environment this is a shape descriptor (box or cylinder).
\end{enumerate}

The parser produces a structured \texttt{ParsedCommand} with fields for \texttt{raw\_text}, \texttt{target\_object}, \texttt{target\_attribute}, \texttt{spatial\_relation}, \texttt{spatial\_reference}, \texttt{location}, and \texttt{action}. The combined attribute + target (e.g., ``blue box'') forms the \texttt{query\_text} used to query the semantic map.

\subsection{Mission Controller}

The \texttt{MissionControllerNode} orchestrates the full autonomous pipeline via a finite state machine utilizing five states: \textsc{idle}, \textsc{exploring}, \textsc{detecting}, \textsc{navigating}, and \textsc{failed}. Upon initialization, the system rests in \textsc{idle} until an explicit text command (e.g., ``start exploration'') is received via the \texttt{/user\_command} topic, at which point the node transitions to \textsc{exploring}. During this phase, Grounding DINO detection runs periodically (every 1.5\,m of travel) alongside the frontier explorer to populate the semantic map.

When an operator issues a specific natural language navigation directive (e.g., ``Go to the green cylinder''), the controller parses the string, queries the stored semantic map for the object's spatial coordinates, and transitions to \textsc{navigating}. These target coordinates are dispatched to the Nav2 framework using the \texttt{/navigate\_to\_pose} action interface, which assumes responsibility for global path planning and local costmap-based obstacle avoidance. Upon task completion (success or failure), the controller returns to \textsc{idle}.

Throughout the mission, the controller subscribes to \texttt{/camera/image\_raw}, \texttt{/depth\_camera/depth/image\_raw}, and \texttt{/odom}, while publishing human-readable telemetry and task outcomes to \texttt{/navigation\_status} and newly detected objects as JSON to \texttt{/semantic\_map/add}. To ensure robustness against memory constraints or system crashes, the mission controller automatically serializes and saves the semantic map to disk after every successful detection cycle.

%-----------------------------------------------------------------------
\section{Implementation}

\subsection{Development Environment}

The complete system is developed and tested on Ubuntu 22.04 LTS with ROS2 Humble Hawksbill, using VSCode as the primary development environment. Gazebo and RViz run natively on the desktop with standard X11 display.

\subsection{Software Dependencies}

The system depends on ROS2 Humble packages (\texttt{turtlebot3}, \texttt{turtlebot3\_gazebo}, \texttt{cartographer\_ros}, \texttt{navigation2}, \texttt{nav2\_bringup}) and Python 3.10 with a dedicated virtual environment containing PyTorch (CPU build), \texttt{transformers} $\geq$4.35, Pillow, OpenCV, SciPy, and spaCy (an optional dependency of the command parser). Grounding DINO model weights are downloaded automatically from HuggingFace on first use. The ROS2 workspace uses \texttt{colcon} for building with \texttt{symlink-install} mode to enable rapid iteration.

\subsection{Parameter Tuning}
\label{sec:params}

Several parameters were empirically tuned for the single-room environment: minimum frontier size $k_{\min} = 3$ cells; goal timeout 60\,s; physical stuck timeout 45\,s; detection interval 1.5\,m; merge distance 0.5\,m; Grounding DINO box threshold 0.30; text threshold 0.25; and a minimum confidence of 0.28 for adding detections to the semantic map. The exploration phase has a configurable time limit (default 900\,s) to prevent indefinite exploration.

%-----------------------------------------------------------------------
\section{Experiments and Results}

\subsection{Experimental Setup}

All experiments were conducted in the Gazebo single-room simulation world described in Section~\ref{sec:arch}. Each experiment begins with the robot spawned at the origin with an empty semantic map. The exploration phase runs for up to 15 minutes or until the frontier explorer reports no remaining frontiers. During exploration, Grounding DINO detection runs every 1.5 meters on CPU (no GPU acceleration). After exploration completes, we evaluate the command parser and navigation pipeline by issuing natural language commands and recording whether the robot successfully navigates to each target object. All results in this section come from a single representative run.

\subsection{Frontier Exploration Performance}

Table~\ref{tab:exploration} summarizes statistics from the representative exploration run. Exploration ended after about 8 minutes, before the 15-minute limit, while parts of the room remained unmapped. The initial implementation used a closest-frontier strategy, which frequently selected frontiers within the Nav2 goal tolerance ($<0.25$\,m), causing immediate ``success'' without robot movement. Switching to the weighted strategy resolved this by favoring larger, more distant frontiers. The adaptive failure recovery was triggered during the run when the robot became stuck near obstacles, allowing it to escape by switching to the farthest-frontier strategy.

\begin{table}[t]
\centering
\caption{Frontier Exploration Statistics (Representative Run)}
\label{tab:exploration}
\begin{tabular}{lc}
\toprule
\textbf{Metric} & \textbf{Value} \\
\midrule
Total navigation goals sent & 16 \\
Goals reached (success) & 12 \\
Goals failed (timeout/stuck) & 4 \\
Goal success rate & 75.0\% \\
Exploration strategy (primary) & Weighted \\
Total exploration duration & $\sim$8\,min \\
Final map coverage (approximate) & $\sim$60\% \\
Semantic map entries stored & 21 \\
\bottomrule
\end{tabular}
\end{table}

\subsection{Object Detection Results}

During the exploration run, Grounding DINO produced 21 detections that were stored in the semantic map. Table~\ref{tab:detection} summarizes the detection results aggregated by object class after label cleaning. The Entries column covers 12 of the 21 stored entries; the remaining 9 are not assigned to a class in this table. Confidence scores varied significantly across object types: the yellow box and red cylinder achieved the highest scores (0.51 each), followed by the green cylinder (0.46). The white cylinder was never correctly detected; its only entry is a spurious match, as discussed below. Its recorded confidence (0.24) is below the thresholds listed in Section~\ref{sec:params}, so it should have been filtered out; we report it as logged. Low color contrast against the light-colored walls may have contributed to the white cylinder being missed.

\begin{table}[t]
\centering
\small
\setlength{\tabcolsep}{4pt}
\caption{Object Detection Results by Class}
\label{tab:detection}
\resizebox{\columnwidth}{!}{
\begin{tabular}{lcccc}
\toprule
\textbf{Object Class} & \textbf{Max Conf.} & \textbf{Entries} & \textbf{Best Det. Pos.} & \textbf{GT Position} \\
\midrule
Yellow box     & 0.51 & 5 & $(2.4,\,-2.8)$ & $(2.5,\,-2.5)$ \\
Red cylinder   & 0.51 & 2 & $(-2.1,\,-0.4)$ & $(-2,\;2)$ \\
Green cylinder & 0.46 & 4 & $(-1.7,\,-2.1)$ & $(-2,\,-2)$ \\
Blue box       & ---  & 0 & ---             & $(2,\;2)$ \\
White cylinder & 0.24 & 1 & $(-3.0,\,-4.4)$ & $(2.5,\;0.5)$ \\
\bottomrule
\end{tabular}
}
\end{table}

Several observations are notable. First, for the three correctly detected objects, the best position estimates deviate from ground truth by 0.3--2.4\,m (yellow box and green cylinder about 0.3\,m, red cylinder about 2.4\,m), due to depth estimation errors at distance, the camera HFOV mismatch noted in Section~\ref{sec:localization}, and accumulated pose error in the TF tree. Second, the blue box at $(2,\,2)$ was not detected because the robot's exploration trajectory, which primarily covered the left half of the room, did not bring it within detection range. Third, the white-cylinder entry at $(-3.0,\,-4.4)$ lies outside the room boundaries and is a spurious match rather than a detection of the real object, so we count the white cylinder as not detected.

A notable challenge is the \emph{multi-object label concatenation problem}. When multiple objects appear in a single camera frame, Grounding DINO sometimes returns a single bounding box with a concatenated label such as ``red cylinder blue box green cylinder yellow box.'' Our label cleaning heuristic matches the raw label against all prompt terms and returns the longest matching term. Despite this, approximately 40\% of the 21 raw detections contained concatenated labels before cleaning. This concatenation also impairs duplicate merging, since detections of the same physical object receive different labels and are stored as separate entries.

\subsection{Language-Guided Navigation}

We evaluated the end-to-end command parsing and navigation pipeline by issuing natural language commands after exploration completed. Table~\ref{tab:navigation} shows the results. The command parser correctly extracted target attributes and object names for both tested commands. The semantic map's substring query successfully matched entries despite the label concatenation issue (e.g., querying ``red cylinder'' matches the stored label ``red cylinder red pillar'' via substring containment). The robot reached the stored map position in both cases. Reaching the map goal is not the same as reaching the object, however: the best red-cylinder estimate lies about 2.4\,m from the true position (Table~\ref{tab:detection}), while the yellow-box estimate lies within about 0.3\,m. 

\begin{table}[t]
\centering
\small
\setlength{\tabcolsep}{4pt}
\caption{Natural Language Command Execution Results (``Reached'' means Nav2 reported success at the stored map position)}
\label{tab:navigation}
\resizebox{\columnwidth}{!}{
\begin{tabular}{lccc}
\toprule
\textbf{Natural Language Command} & \textbf{Parsed Target} & \textbf{Map Hit} & \textbf{Result} \\
\midrule
Go to the yellow box   & yellow box   & Yes (0.51) & \checkmark~reached \\
Go to the red cylinder & red cylinder & Yes (0.51) & \checkmark~reached \\
\bottomrule
\end{tabular}
}
\end{table}

\begin{figure}[t]
    \centering
    \includegraphics[width=0.5\columnwidth]{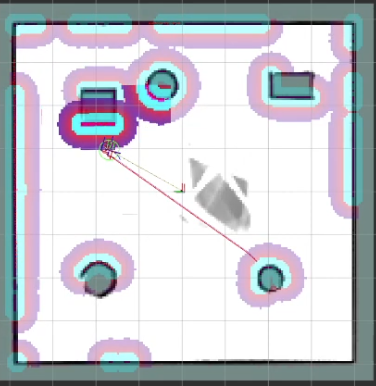}
    \caption{Nav2 executing a natural language navigation command. The green line is the global path from the NavFn planner, the red curve is the local trajectory from the DWB controller, and the arrow marker indicates the goal pose derived from the semantic map query.}
    \label{fig:navexec}
\end{figure}

%-----------------------------------------------------------------------
\section{Discussion}

\subsection{Key Findings}

Our experiments suggest three findings. First, a complete vision-language navigation pipeline is achievable on commodity CPU hardware without GPU acceleration, though with inference latency of 5--10 seconds per image. Second, the weighted frontier strategy with adaptive failure recovery mitigates a critical failure mode observed during development: the closest-frontier strategy frequently selected goals within Nav2's 0.25\,m goal tolerance, causing the robot to report ``success'' without moving. The weighted strategy, by balancing frontier size against distance, avoids this degenerate case. Third, regex-based command parsing was sufficient for the navigation commands we tested, and avoids the latency and deployment complexity of LLM-based parsers.

\subsection{Limitations}

Multi-object label concatenation is the most significant issue in our current detection pipeline. When many objects appear in one frame, Grounding DINO returns merged labels, causing incorrect 3D localization and preventing proper duplicate merging. Depth-based 3D localization error of 0.3--2.4\,m is significant relative to the object sizes, and is compounded by a mismatch between the Gazebo camera HFOV and the intrinsics used in the code. Exploration covered only about 60\% of the room, which left one object undetected. Finally, all results come from a single exploration run and two navigation commands; they demonstrate feasibility but do not provide statistically meaningful performance estimates.

\subsection{Future Work}

Five directions would significantly improve the system. First, a principled solution to multi-object label concatenation could be implemented by prompting Grounding DINO with one object category at a time and aggregating results, at the cost of additional inference calls. Second, GPU acceleration via CUDA would bring detection to real-time rates suitable for physical robot deployment. Third, hierarchical semantic maps that organize objects by room would enable room-level navigation commands such as ``go to the kitchen table.'' Fourth, replacing the regex parser with a small on-device language model would enable more complex spatial and compositional queries. Fifth, extension to a multi-room house environment would test the system's scalability beyond the single-room environment.

%-----------------------------------------------------------------------
\section{Conclusion}

We presented a complete vision-language grounded navigation system integrating frontier-based exploration, Grounding DINO open-vocabulary detection, depth-based 3D localization, semantic mapping with duplicate merging, and regex-based natural language command parsing in a unified ROS2 Humble pipeline. Experiments in a Gazebo single-room simulation containing five geometric objects demonstrated autonomous exploration covering approximately 60\% of the room in about 8 minutes, detection of three of the five objects with confidence scores up to 0.51, and successful execution of both natural language navigation commands tested. The weighted frontier selection with adaptive failure recovery prevented the robot from becoming permanently stuck during the run. The system operates entirely on CPU-only commodity hardware without GPU acceleration. The key remaining challenges are reducing multi-object label confusion from Grounding DINO, correcting the camera intrinsics mismatch to improve 3D localization accuracy, improving exploration coverage, and scaling the system to multi-room environments.

%-----------------------------------------------------------------------

\end{document}